\documentclass{article}
\usepackage{spconf}
\usepackage{amsmath}
\usepackage{graphicx}
\usepackage{xcolor}
\usepackage{tabu}
\usepackage[utf8]{inputenc}
\usepackage{enumitem}

\usepackage{caption}
\usepackage{subcaption}

\usepackage[hidelinks]{hyperref}
\urldef{\supplementary}\url{https://youtu.be/fgmrTlJJRe4}
\usepackage[sort&compress,noabbrev,capitalise,nameinlink]{cleveref}

\newcommand\Tstrut{\rule{0pt}{2.6ex}}         
\newcommand\Bstrut{\rule[-0.8ex]{0pt}{0pt}}   

\let\OLDthebibliography\thebibliography
\renewcommand\thebibliography[1]{
  \OLDthebibliography{#1}
  \setlength{\parskip}{0pt}
  \setlength{\itemsep}{0pt plus 0.3ex}
}

\title{FlowFields++:\\Accurate Optical Flow Correspondences Meet Robust Interpolation}
\name{René Schuster, Christian Bailer, Oliver Wasenmüller, Didier Stricker}
\address{DFKI -- German Research Center for Artificial Intelligence\\
\texttt{firstname.lastname@dfki.de}}
\begin{document}
%
\maketitle
\begin{abstract}
Optical Flow algorithms are of high importance for many applications. Recently, the Flow Field algorithm and its modifications have shown remarkable results, as they have been evaluated with top accuracy on different data sets. In our analysis of the algorithm we have found that it produces accurate sparse matches, but there is room for improvement in the interpolation. Thus, we propose in this paper FlowFields++, where we combine the accurate matches of Flow Fields with a robust interpolation. In addition, we propose improved variational optimization as post-processing. Our new algorithm is evaluated on the challenging KITTI and MPI Sintel data sets with public top results on both benchmarks. 
\end{abstract}

\section{Introduction} \label{sec:intro}
One core component of machine vision in many domains is the estimation of dense optical flow, e.g. as input for other algorithms such as 3D reconstruction or odometry, in autonomous driving and robot navigation as perception of the motion of the environment, and many more.
More and more vision-based applications require increasing performance of their underlying algorithms. That includes faster run time to enable real time applications, higher accuracy to distinguish from competitors, or increased robustness under challenging environmental conditions to improve the reliability.
The need for steadily improving optical flow estimation has led to impressive results in research and industry regarding the performance and diversity of approaches.

\begin{figure}[t]
	\centering
	\begin{subfigure}[c]{0.48\columnwidth}
		\includegraphics[width=1\textwidth]{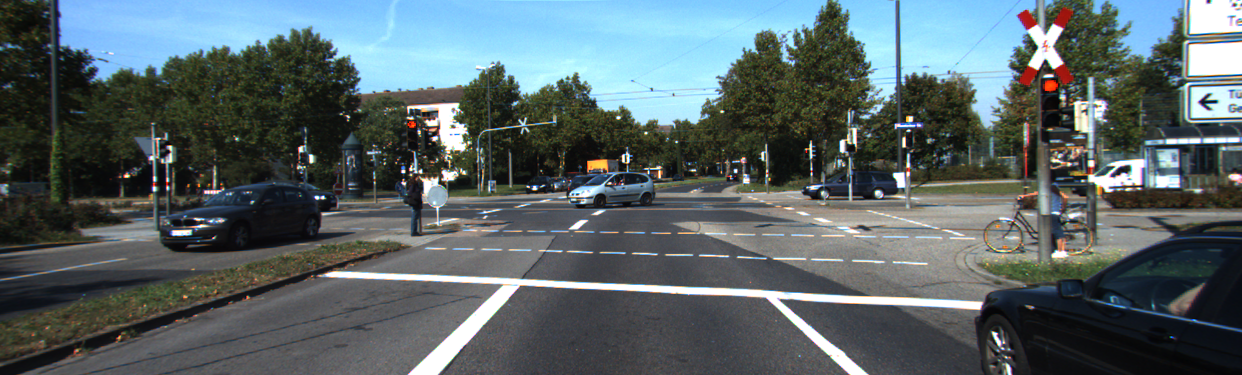}%
		\subcaption{Reference Image}
	\end{subfigure}
	\begin{subfigure}[c]{0.48\columnwidth}
		\includegraphics[width=1\textwidth]{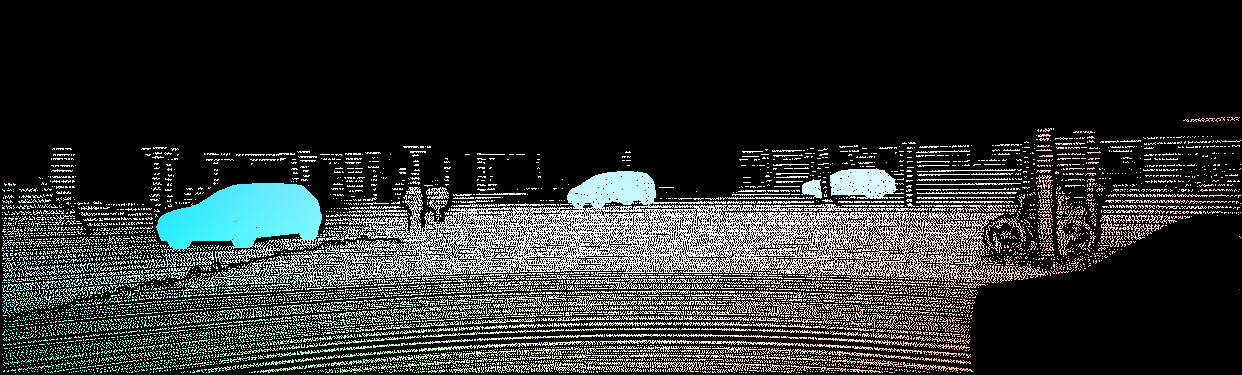}%
		\subcaption{Ground Truth}
	\end{subfigure}
	\par\medskip
	\begin{subfigure}[c]{0.48\columnwidth}
		\includegraphics[width=1\textwidth]{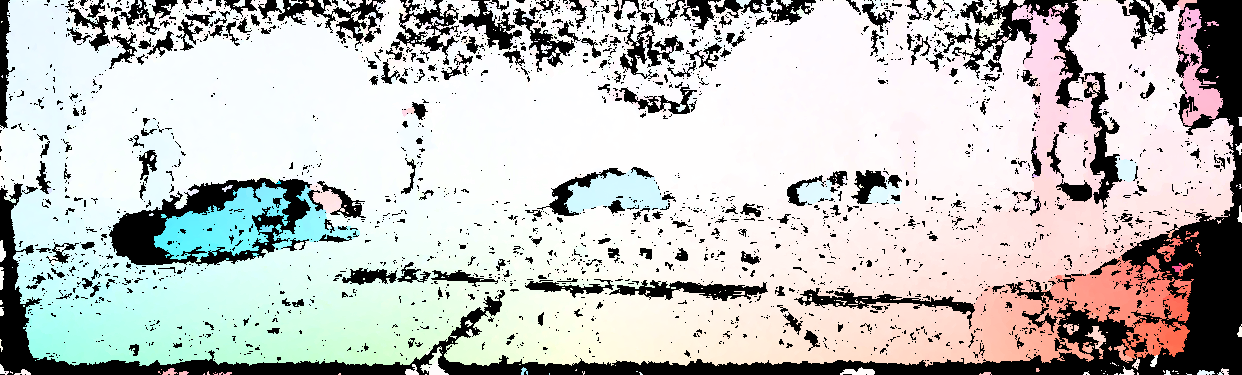}%
		\subcaption{Sparse Matching}
	\end{subfigure}
	\begin{subfigure}[c]{0.48\columnwidth}
		\includegraphics[width=1\textwidth]{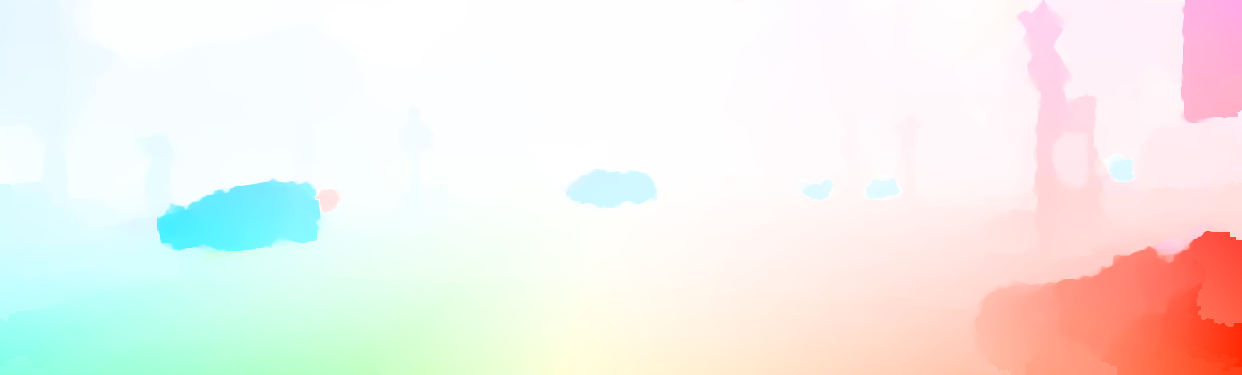}%
		\subcaption{Dense Interpolation}
	\end{subfigure}
	\caption{FlowFields++: Accurate matches get robustly interpolated for dense optical flow.}
	\label{fig:title}
\end{figure}

In this work, we push the limits further by creating an enhanced optical flow algorithm that performs superior on different data sets and is not dedicated to a single data set only. Towards this end, we have identified Flow Fields \cite{bailer2015flow} as a very versatile state-of-the-art matching approach that we combine with robust interpolation (cf. \cref{fig:title}). The advantages of each separate concept shall surpass the weaknesses of their respective counterpart to boost the overall performance way beyond each individual algorithm.
In detail, our contributions are the following:
\begin{itemize}[noitemsep,topsep=1pt,label={\tiny\raisebox{0.75ex}{\textbullet}},leftmargin=*]
	\item Novel combination of accurate matching with robust interpolation.
	\item Improved variational optimization for optical flow in a dual-frame setting.
	\item Thorough evaluation\footnote{\supplementary} on two different challenging data sets to verify versatility and accuracy.
\end{itemize}

\section{Related Work} \label{sec:relwork}
Flow Fields \cite{bailer2015flow} can be considered as the basis of our method. It was among the first approaches to achieve top performance across multiple data sets and has been refined several times since its publication. Flow Fields+ \cite{bailer2017optical} improved the algorithm by more sophisticated matching, FlowFieldsCNN \cite{bailer2017cnn} used deep learning to evaluate the matching cost, and ProbFlowFields \cite{wannenwetsch2017probflow} improved the results by jointly estimating optical flow and a certainty measure. Our approach shares the basic concepts with the mentioned methods. That is, we also perform dense matching by multi-scale propagation and random search, followed by outlier rejection, post processed by an interpolation mechanism. However, the novel FlowFields++ differs from those approaches by combining the matching accuracy of Flow Fields with robust interpolation.

Dense interpolation has become a very popular post processing step for many applications ever since the publication of the first successful interpolation method EPICFlow \cite{revaud2015epic}. It was used by Flow Fields and many other matching methods to produce dense results. InterpoNet \cite{zweig2017interponet} tried to solve the task of optical flow interpolation with a neural network that showed improvements over EPICFlow depending on input matches and data set. Independent of these factors are the advantages of RICFlow \cite{hu2017robust} over EPICFlow. The basic idea of edge preserving interpolation is complemented by increased robustness in the computation of the piece-wise interpolation models. We will exploit this robustness for our approach and extend it further by improved edge detectors and adjusted variational refinement.

In recent eras of deep learning, there are also approaches that use convolutional neural networks (CNNs) to aid or solve the task of optical flow estimation. Some try to compute optical flow in and end-to-end manner \cite{dosovitskiy2015flownet,ilg2017flownet,meister2018unflow}, others use neural networks to compute a semantic segmentation as additional input \cite{sevilla2016optical,bai2016exploiting,hur2016joint}, and some use deep learning for matching cost computation \cite{bailer2017cnn,revaud2016deepmatching,guney2016deep}. Of course, all deep learning approaches require a lot of proper training data and none yet has showed to generalize well across different data sets without retraining or tuning. Our approach maintains its versatility by avoiding deep learning.

\begin{figure}[t]
	\begin{center}
		\begin{subfigure}[c]{0.32\columnwidth}
			\includegraphics[width=1\textwidth]{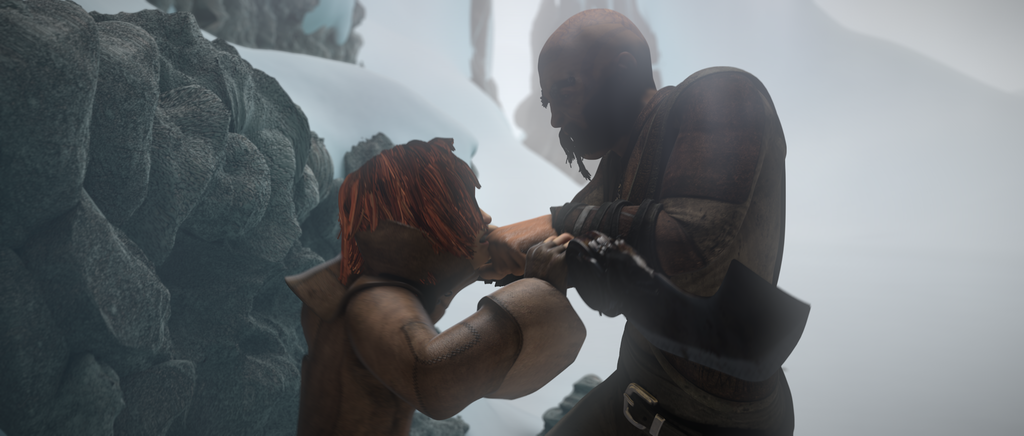}
			\subcaption{Input}
		\end{subfigure}
		\begin{subfigure}[c]{0.32\columnwidth}
			\includegraphics[width=1\textwidth]{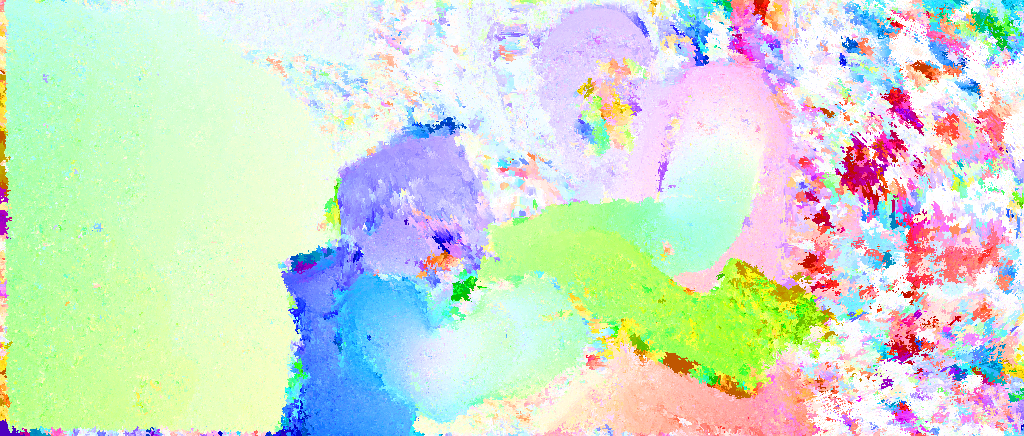}
			\subcaption{Matching}
		\end{subfigure}
		\begin{subfigure}[c]{0.32\columnwidth}
			\includegraphics[width=1\textwidth]{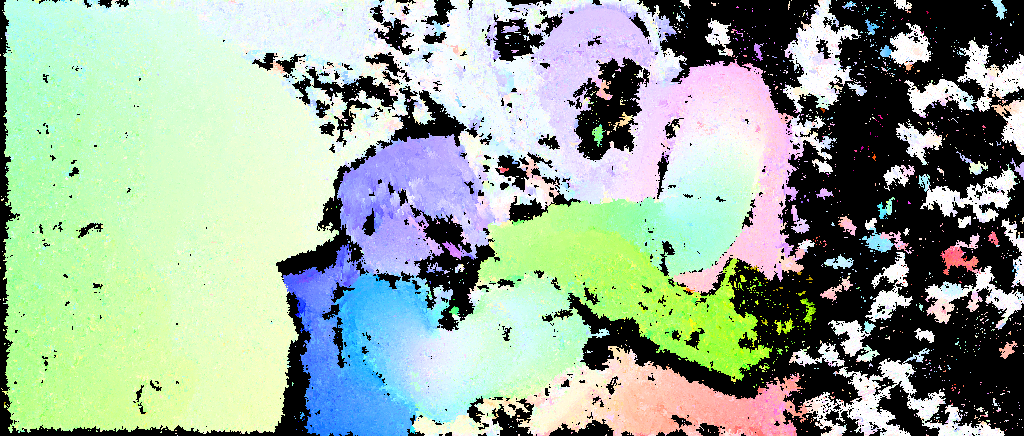}
			\subcaption{Filtering}
		\end{subfigure}
		\par\medskip
		\begin{subfigure}[c]{0.32\columnwidth}
			\includegraphics[width=1\textwidth]{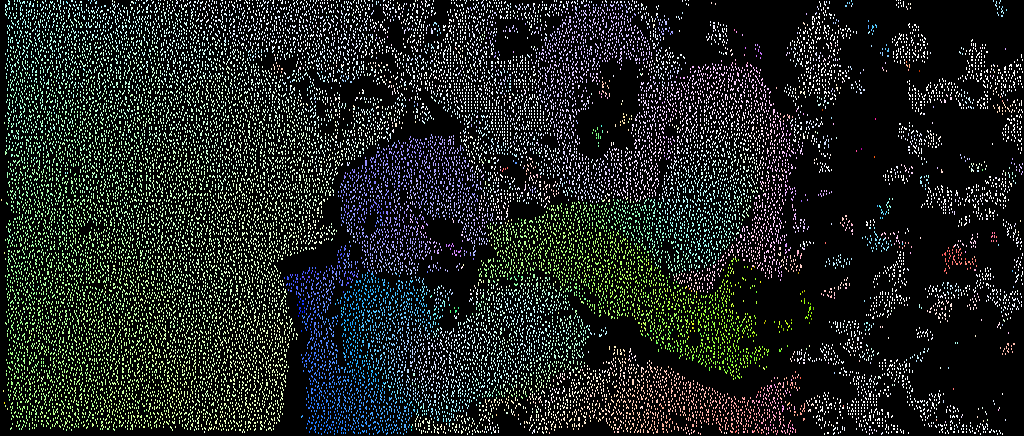}
			\subcaption{Sparsification}
		\end{subfigure}
		\begin{subfigure}[c]{0.32\columnwidth}
			\includegraphics[width=1\textwidth]{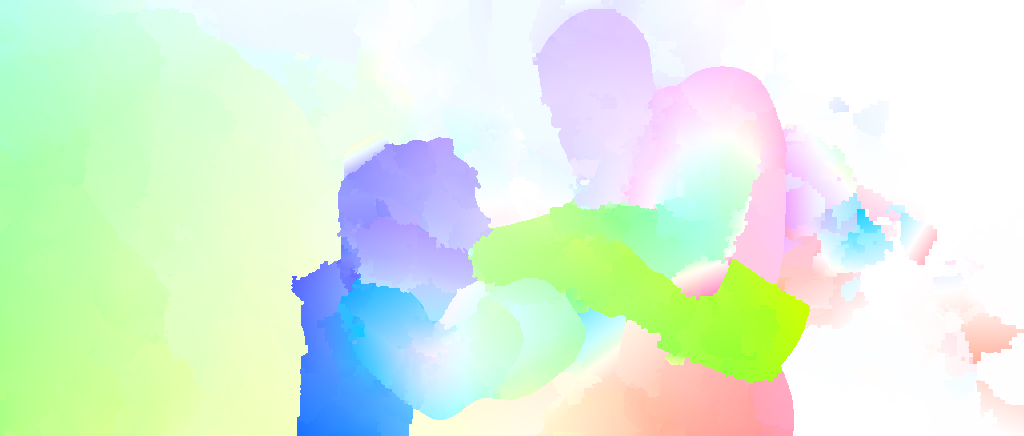}
			\subcaption{Interpolation}
		\end{subfigure}
		\begin{subfigure}[c]{0.32\columnwidth}
			\includegraphics[width=1\textwidth]{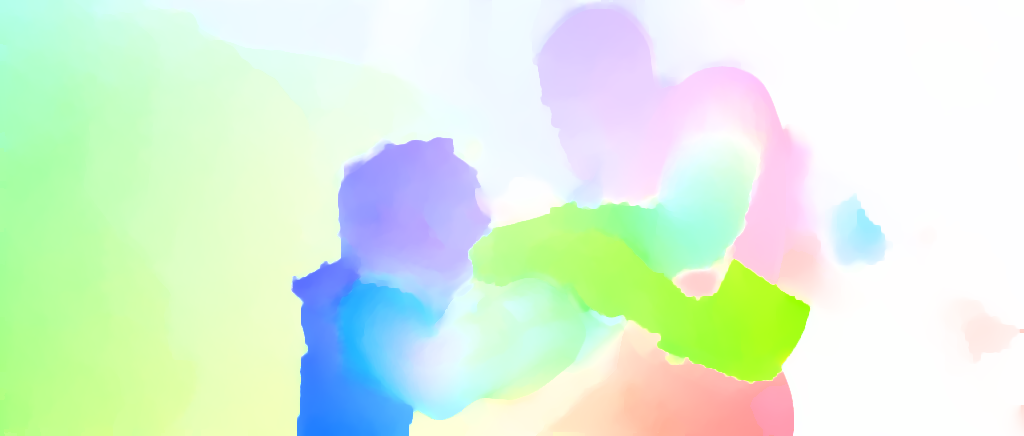}
			\subcaption{Optimization}
		\end{subfigure}
	\end{center}
	\captionsetup{aboveskip=0pt}
	\caption{Visualization of the FlowFields++ pipeline.}
	\label{fig:overview}
\end{figure}

\section{Method} \label{sec:method}
As mentioned before, our method consists mainly of two parts. First, the computation of sparse, accurate optical flow correspondences, and secondly, the robust interpolation to a dense flow field (see \cref{fig:title}). As others before, we apply a coarse-to-fine matching approach using spatial propagation and random search \cite{bailer2015flow}. A strong multi-stage filtering step is reliably removing most of the falsely matched correspondences. Robust, edge-aware interpolation fills up the filtered gaps efficiently \cite{hu2017robust}. Sharp edges are obtained via a random forest trained on semantic boundaries \cite{dollar2013sed,schuster2018sceneflowfields}. Final refinement is done via an adjusted variational framework \cite{brox2004high}. The separate steps are visualized in \cref{fig:overview}.

\subsection{Matches} \label{sec:matches}
We define the matching cost of corresponding pixels as a distance between their feature descriptors. For our experiments we consider two terms. Either, the Hamming distance of the binary patterns created by the Census transform \cite{zabih1994census} in a $7\times 7$ patch for all channels of the CIELab color space, or the Euclidean distance of SIFT features \cite{lowe1999sift}.
Sub-pixel locations are interpolated bilinearly in both cases. 

For initialization on lowest resolution, Walsh-Hadamard feature descriptors are matched using kD-tress \cite{hel2005real}. Those initial matches are propagated into all four quadrants for $i=12$ iterations at each scale. We use sub-scales and sub-sub-scales as in \cite{bailer2017optical}. Additionally, after each of the last three propagation steps of one iteration, a random search is performed where a small random offset is added to each flow component.
To reject outliers introduced during the matching process, we perform two consistency checks and region-based filtering during sparsification. Each consistency check compares the estimated optical flow with an inverse flow field computed with different matching parameters and removes the estimated flow if the consistency error exceeds a consistency threshold $\epsilon$. Finally after filtering, we sparsify the remaining matches further by only selecting matches with the lowest consistency error in each $3\times 3$ block that contains at least $s$ matches. The sparsification helps to increase the spatial support during interpolation and has a positive effect on the run time.
Useful implementation details are given in \cite{bailer2017optical}. 

\subsection{Interpolation} \label{sec:interpolation}
Our interpolation strategy has two core properties: Edge-awareness and robustness. Following the ideas of \cite{revaud2015epic,hu2017robust} we compute local neighborhoods with respect to a geodesic distance that is based on image edges. To simulate euclidean distances, an offset of $0.002$ is added to the edge maps. Contrary to the original work in this field, we do not rely on basic image edges but rather train a random forest on semantic boundaries using the appraoch of \cite{schuster2018sceneflowfields}. It was recently shown that this kind of edges are superior to the standard SED \cite{dollar2013sed} model. Further we use the approximation of \cite{hu2017robust} by segmenting the images into super-pixels to find an optimal consensus set out of a local super-pixel neighborhood to compute the interpolation model. Since random sampling alone would be too inefficient for model estimation, this approach performs spatial propagation of already estimated models. The interpolation model is an affine 2D transformation of 6 unknowns that is used to transform all pixels where optical flow values are missing. Compared to previous methods like EPICFlow \cite{revaud2015epic}, this strategy is much more robust because model estimation is not based on all matches of the local neighborhood. This way, the additional inlier selection can successfully reject remaining outliers in the matches.

\begin{table}[t]
	\caption{Results on \textbf{KITTI 2015}. We compare average percentage of outliers on background \textit{Fl-bg}, foreground \textit{Fl-fg}, and all \textit{Fl-all} pixels. The best published monocular methods are listed. Multi-frame and deep learning appraoches are given in gray. Run times on a GPU are in parenthesis.} \label{tab:kitti}
	\centering
	\resizebox{\columnwidth}{!}{\begin{tabu}{c | r | r | r | c }
		{\bf Method} & {\bf Fl-bg} & {\bf Fl-fg} & {\bf Fl-all} & {\bf Time} \Bstrut\\
		\hline
		MirrorFlow \cite{hur2017mirrorflow} & 8.93 & 17.07 & 10.29 & 660 s \Tstrut\\
		\rowfont{\color{gray}}
		SDF \cite{bai2016exploiting} & 8.61 & 23.01 & 11.01 & --- \\
		\rowfont{\color{gray}}
		UnFlow \cite{meister2018unflow} & 10.15 & 15.93 & 11.11 & (0.12 s) \\
		\rowfont{\color{gray}}
		CNNF+PMBP \cite{zhang2018fundamental} & 10.08 & 18.56 & 11.49 & 2700 s \\
		\rowfont{\color{gray}}
		MR-Flow \cite{wulff2017optical} & 10.13 & 22.51 & 12.19 & 480 s \\
		\rowfont{\color{gray}}
		DCFlow \cite{xu2017accurate} & 13.10 & 23.70 & 14.86 & (8.6 s) \\
		\textbf{FlowFields++ (ours)} & \textbf{14.82} & \textbf{17.77} & \textbf{15.31} & \textbf{29 s} \\
		\rowfont{\color{gray}}
		SOF \cite{sevilla2016optical} & 14.63 & 22.83 & 15.99 & 360 s \\
		\rowfont{\color{gray}}
		JFS \cite{hur2016joint} & 15.90 & 19.31 & 16.47 & 780 s \\
		DF+OIR \cite{maurer2017order} & 15.11 & 23.45 & 16.50 & 180 s \\
		\rowfont{\color{gray}}
		ImpPB+SPCI \cite{schuster2017optical} & 17.25 & 20.44 & 17.78 & (60 s) \\
		\rowfont{\color{gray}}
		FlowFieldCNN \cite{bailer2017cnn} & 18.33 & 20.42 & 18.68 & (23 s) \\
		RicFlow \cite{hu2017robust} & 18.73 & 19.09 & 18.79 & 5 s \\
		FlowFields+ \cite{bailer2017optical} & 19.51 & 21.26 & 19.80 & 28 s \\
	\end{tabu}}
\end{table}

\subsection{Variational Optimization} \label{sec:variational}
\begin{figure*}[t]
	\begin{center}
		\hspace{0.05\textwidth}
		\begin{subfigure}[c]{0.42\textwidth}
			\centering
			First Image%
			\vspace{0.25mm}%
			\end{subfigure}
		\begin{subfigure}[c]{0.42\textwidth}
			\centering
			Second Image%
			\vspace{0.25mm}%
		\end{subfigure}
		\\%
		\hspace{0.05\textwidth}
		\begin{subfigure}[c]{0.42\textwidth}
			\includegraphics[width=1\textwidth]{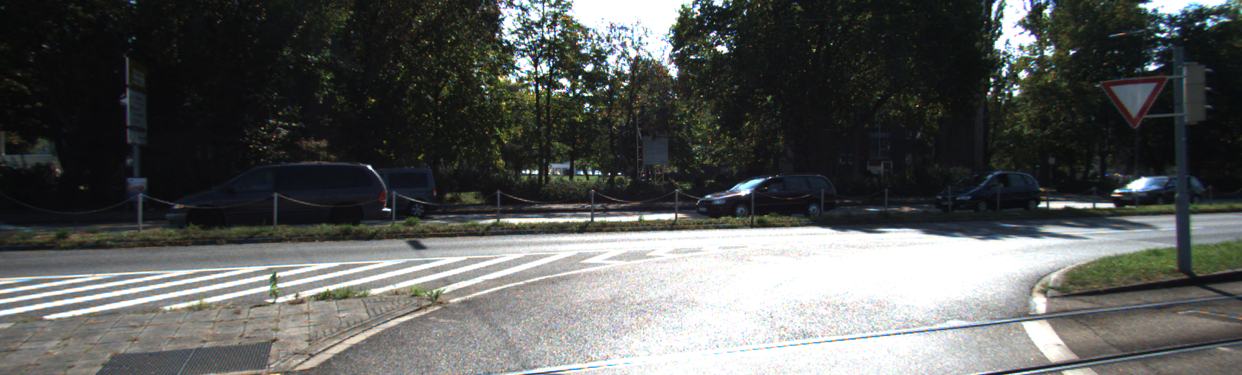}%
			\vspace{1mm}%
		\end{subfigure}
		\begin{subfigure}[c]{0.42\textwidth}
			\includegraphics[width=1\textwidth]{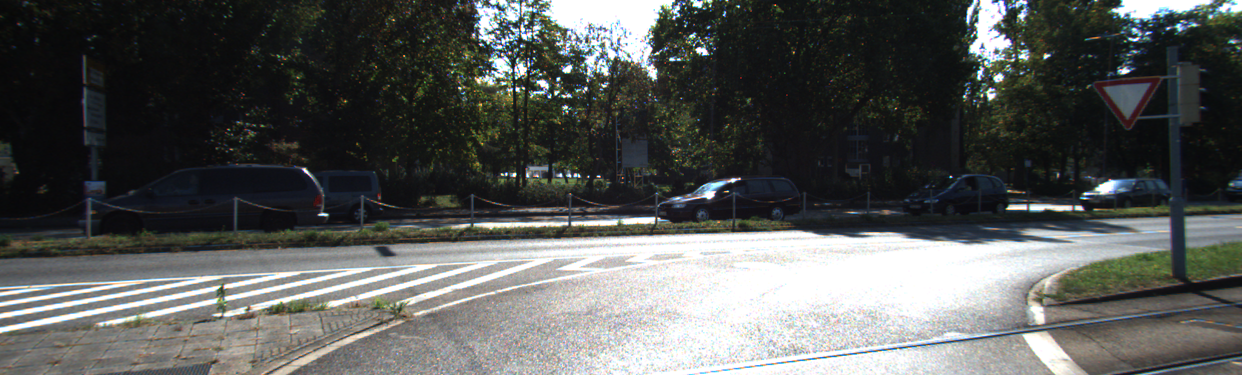}%
			\vspace{1mm}%
		\end{subfigure}
		\\%
		\hspace{0.05\textwidth}
		\begin{subfigure}[c]{0.42\textwidth}
			\centering
			Flow Estimates%
			\vspace{0.25mm}%
			\end{subfigure}
		\begin{subfigure}[c]{0.42\textwidth}
			\centering
			Error Maps%
			\vspace{0.25mm}%
		\end{subfigure}
		\\%
		\begin{subfigure}[c][5mm][c]{0.05\textwidth}
			\rotatebox[origin=c]{90}{\begin{tabular}[b]{@{}c@{}}MirrorFlow\\\cite{hur2017mirrorflow}\end{tabular}}%
		\end{subfigure}
		\begin{subfigure}[c]{0.42\textwidth}
			\includegraphics[width=1\textwidth]{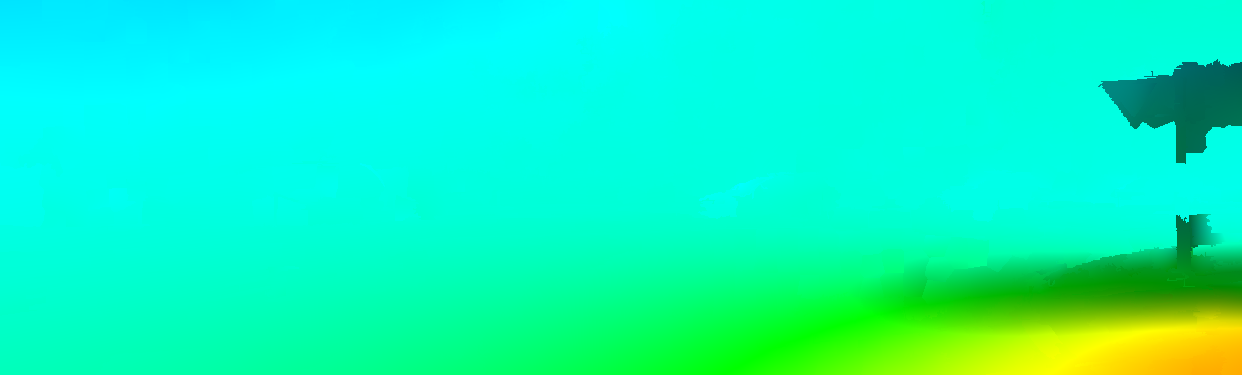}%
			\vspace{0.25mm}%
		\end{subfigure}
		\begin{subfigure}[c]{0.42\textwidth}
			\includegraphics[width=1\textwidth]{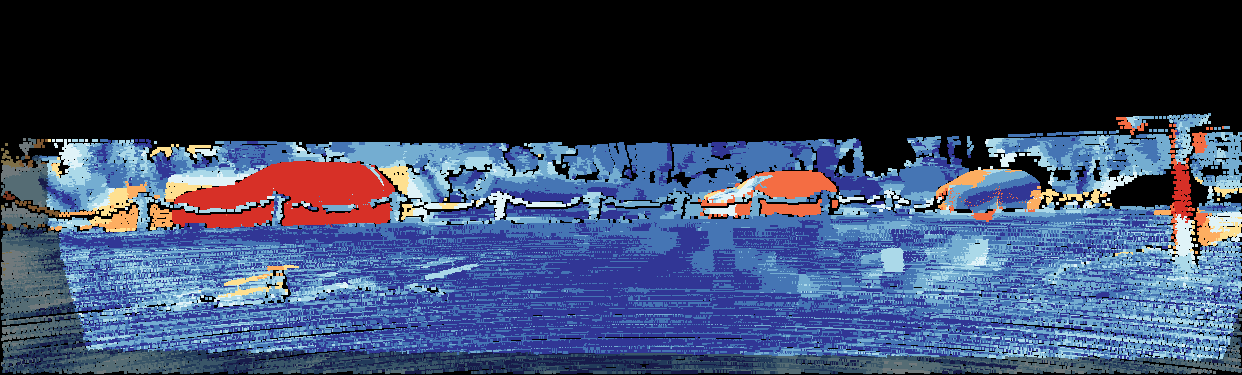}%
			\vspace{0.25mm}%
		\end{subfigure}
		\\%
		\begin{subfigure}[c][5mm][c]{0.05\textwidth}
			\rotatebox[origin=c]{90}{\begin{tabular}[b]{@{}c@{}}FlowFields+\\{\cite{bailer2017optical}}\end{tabular}}%
		\end{subfigure}
		\begin{subfigure}[c]{0.42\textwidth}
			\includegraphics[width=1\textwidth]{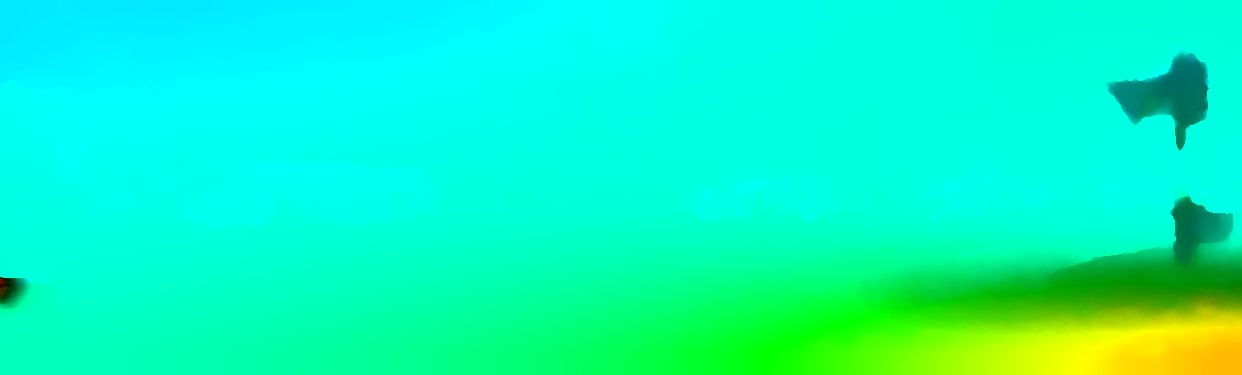}%
			\vspace{0.25mm}%
		\end{subfigure}
		\begin{subfigure}[c]{0.42\textwidth}
			\includegraphics[width=1\textwidth]{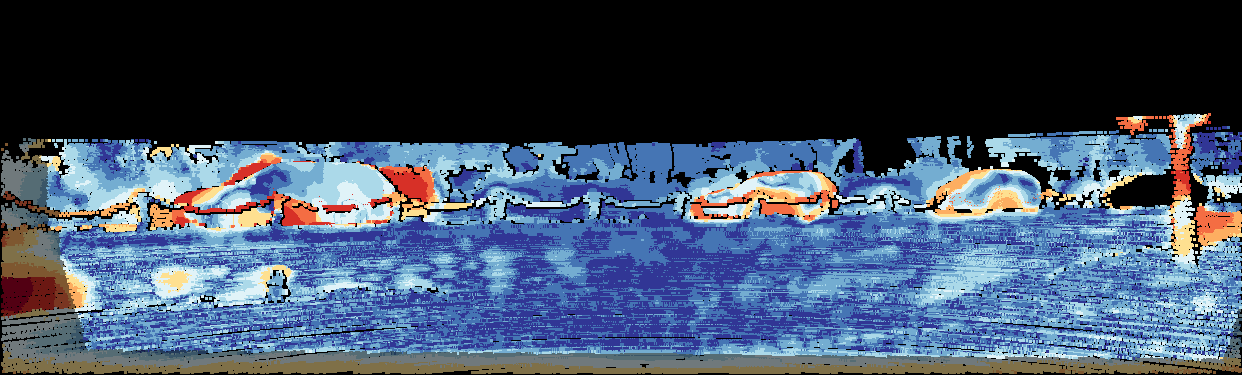}%
			\vspace{0.25mm}%
		\end{subfigure}
		\\%
		\begin{subfigure}[c][5mm][c]{0.05\textwidth}
			\rotatebox[origin=c]{90}{\bf \begin{tabular}[b]{@{}c@{}}FlowFields++\\(ours)\end{tabular}}%
		\end{subfigure}
		\begin{subfigure}[c]{0.42\textwidth}
			\includegraphics[width=1\textwidth]{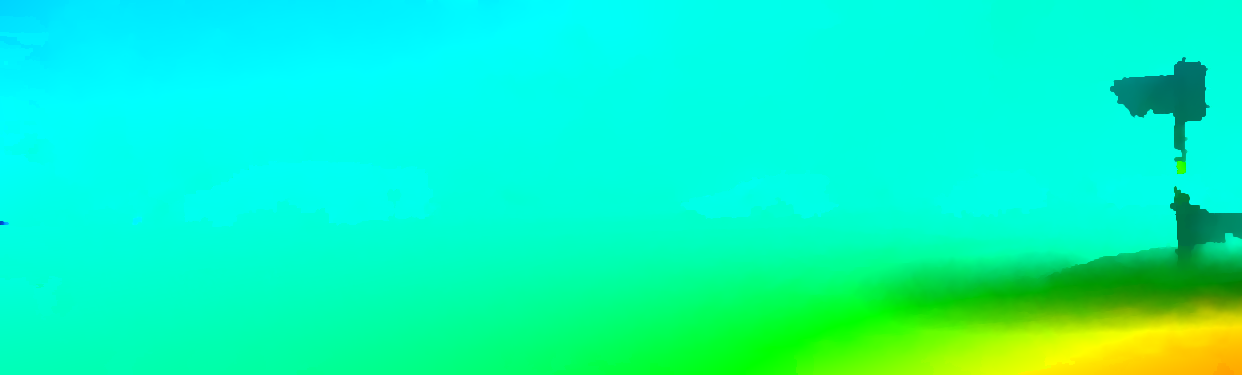}%
			\vspace{0.25mm}%
		\end{subfigure}
		\begin{subfigure}[c]{0.42\textwidth}
			\includegraphics[width=1\textwidth]{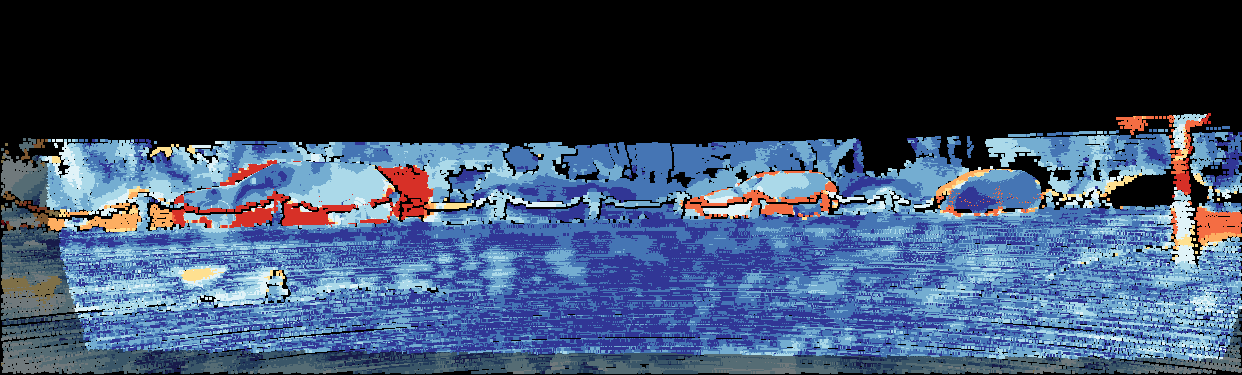}%
			\vspace{0.25mm}%
		\end{subfigure}
		\\%
	\end{center}
	\caption{Visual results on the test data of KITTI. We show the estimated optical flow and the according error maps for MirrorFlow, FlowFields+, and our FlowFields++. Contrary to FlowFields+, our approach estimates out-of-bounds regions correctly.} 
	\label{fig:results}
\end{figure*}

The densely interpolated flow field is then used as initialization for variational refinement. We use the framework of Brox et al. \cite{brox2004high} with two important adjustments. First, the coarse-to-fine pyramid steps are not required because our dense flow field is already a very good estimate on full resolution. Secondly, we do not optimize the optical flow where it would leave the image domain. This is very important because the variational energy at those regions is solely defined by the smoothness term. Thus, the optimization process tends towards constant optical flow which is often inappropriate. By leaving these areas out, we rather rely on our already precise interpolation instead of taking the high risk of oversmoothing. Evidence for this decision is given by the out-of-bounds regions of the KITTI data set in \cref{fig:results}.

\begin{table}[t]
	\caption{Results on \textbf{MPI Sintel}. We give the average end-point error (EPE) on the \textit{final} rendering pass for \textit{matched} and \textit{unmatched} regions and \textit{all} pixels. Only the best variant of each published method is listed. Multi-frame and deep learning appraoches are given in gray.} \label{tab:sintel}
	\centering
	\resizebox{\columnwidth}{!}{\begin{tabu}{c | c | c | c }
		{\bf Method} & {\bf \begin{tabular}[b]{@{}c@{}}EPE\\all\end{tabular}} & {\bf \begin{tabular}[b]{@{}c@{}}EPE\\matched\end{tabular}} & {\bf \begin{tabular}[b]{@{}c@{}}EPE\\unmatched\end{tabular}} \Bstrut\\
		\hline
		\rowfont{\color{gray}}
		DCFlow \cite{xu2017accurate} & 5.119 & 2.283 & 28.228 \Tstrut\\
		\rowfont{\color{gray}}
		FlowFieldsCNN \cite{bailer2017cnn} & 5.363 & 2.303 & 30.313 \\
		\rowfont{\color{gray}}
		MR-Flow \cite{wulff2017optical} & 5.376 & 2.818 & 26.235 \\
		S2F-IF \cite{yang2017slow} & 5.417 & 2.549 & 28.795 \\
		\textbf{FlowFields++ (ours)} & \textbf{5.486} & \textbf{2.614} & \textbf{28.900} \\	
		\rowfont{\color{gray}}
		InterpoNet \cite{zweig2017interponet} & 5.535 & 2.372 & 31.296 \\
		RicFlow \cite{hu2017robust} & 5.620 & 2.765 & 28.907 \\
		ProbFlowFields \cite{wannenwetsch2017probflow} & 5.696 & 2.545 & 31.371 \\
		FlowFields+ \cite{bailer2017optical} & 5.707 & 2.684 & 30.642 \\
		\rowfont{\color{gray}}
		DeepDiscreteFlow \cite{guney2016deep} & 5.728 & 2.623 & 31.042 \\
		\rowfont{\color{gray}}
		FlowNet2 \cite{ilg2017flownet} & 5.739 & 2.752 & 30.108 \\
		FlowFields \cite{bailer2015flow} & 5.810 & 2.621 & 31.799 \\	
	\end{tabu}}
\end{table}

\section{Results} \label{sec:results}

We evaluate our method on the popular optical flow benchmarks KITTI \cite{menze2015object} and MPI Sintel \cite{butler2012sintel} and show that we belong to the few methods that achieve top performance on both. This again confirms that our generic approach is not restricted to any data domain or setup.

\subsection{Parameter Selection} \label{sec:parameters}
Unless explicitly stated here, we use the same values for both data sets. As matching features, we use SIFT for KITTI and Census for Sintel as described in \cref{sec:matches}. We use different random forests, i.e on KITTI we estimate semantic boundaries as described in \cref{sec:interpolation} and on Sintel we use the default model of \cite{dollar2013sed}. Other variations are given in \cref{tab:parameters}.
All other parameters apply to both data sets and are either given in the previous sections or are the default values of the respective framework. 

\begin{table}[h]
	\caption{Parameters for KITTI and MPI Sintel.}
	\label{tab:parameters}
	\centering
	\begin{tabular}{c | c | c}
		{\bf Parameter} & {\bf KITTI} & {\bf Sintel} \Bstrut\\
		\hline
		Consistency threshold $\epsilon$ & 1 & 7 \Tstrut\\
		Minimum matches $s$ & 7 & 4 \\
		Super-pixel size & 20 & 50 \\
		Local neighborhood size & 150 & 200 \\
		Variational iterations & 2 & 5 \\
	\end{tabular}
\end{table}

\subsection{KITTI 2015} \label{sec:results_kitti}
The first data set on which we evaluate our approach is KITTI \cite{menze2015object}. It consists of traffic scenarios and provides sparse ground truth. The data set contains large displacements and challenging lightning conditions. 

For a visual impression, we give exemplary results of our method compared to others in \cref{fig:results}. Worth highlighting is the impact of our adjusted variational refinement described in \cref{sec:variational}. Compared to Flow Fields+ that uses the variational optimization of EPICFlow, our appraoch estimates the out-of-bounds regions correctly. This can be seen in the shaded regions of the error maps in \cref{fig:results}.

For a qualitative comparison, we show the top performing methods on KITTI in \cref{tab:kitti}. Only three non-learning-based methods perform better than our approach. However, MirrorFlow \cite{hur2017mirrorflow} is not competitive on other data sets, MR-Flow \cite{wulff2017optical} is using more than two frames which resolves ambiguities in invisble parts of the reference image, and DCFlow \cite{xu2017accurate} performs much worse on moving foreground regions that are of high importance in motion estimation. Compared to Flow Fields+ \cite{bailer2017optical} that shares our matching strategy and to RICFlow \cite{hu2017robust} that also uses robust interpolation, we could reduce the average amount of outliers by about 20 \%.

\subsection{MPI Sintel} \label{sec:sintel}
In addition, we use MPI Sintel \cite{butler2012sintel} for evaluation. This data set provides dense ground truth for synthetically rendered sequences. It typically consists of large displacements with fine details, highly non-rigid motions of close-up deformable characters, and challenging image quality due to motion blur and atmospheric effects in the \textit{final} rendering pass (cf. \cref{fig:overview}).
This characteristics are often the reason why algorithms that perform good on KITTI tend to perform much worse on Sintel. Our FlowFields++ belongs to the rare exceptions along with DCFlow \cite{xu2017accurate} and MR-Flow \cite{wulff2017optical}. As mentioned before MR-Flow is using multiple frames which is the reason for its exceptional performance in \textit{unmatched} regions, but in \textit{matched} areas it performance considerably worse than all other listed methods.

\cref{tab:sintel} lists the highest ranked methods on Sintel. Noticeably few methods are also listed in the KITTI ranking. Especially deep learning approaches are rarely represented with top performance on both data sets, besides FlowFieldsCNN \cite{bailer2017cnn}. Again, only three non-learning approaches are ranked higher than FlowFields++ while only one of those is a dual-frame method that is also represented on KITTI. On the \textit{clean} rendering pass FlowFields++ is even the highest ranked dual-frame method.

\section{Conclusion} \label{sec:conclusion}
We have presented FlowFields++, a generic optical flow algorithm by combining highly accurate matching and robust, edge-preserving interpolation. The performance of the approach was evaluated on two diverse public data sets. In a joint ranking of both data sets, we claim to be the second best method among all dual-frame methods (even including deep learning approaches) after DCFlow \cite{xu2017accurate} that we still outperform in foreground regions on KITTI.

\bibliographystyle{IEEEbib}
\bibliography{bib}

\end{document}